# Turing Test 2.0: The General Intelligence Threshold

Georgios Mappouras
george.mappouras@gmail.com

**Abstract**

*With the rise of artificial intelligence (A.I.) and large language models like ChatGPT, a new race for achieving artificial general intelligence (A.G.I.) has started. While many speculate how and when A.I. will achieve A.G.I., there is no clear agreement on how A.G.I. can be detected in A.I. models, even when popular tools like the Turing test (and its modern variations) are used to measure their intelligence. In this work, we discuss why traditional methods like the Turing test do not suffice for measuring or detecting A.G.I. and provide a new, practical method that can be used to decide if a system (computer or any other) has reached or surpassed A.G.I. To achieve this, we make two new contributions. First, we present a clear definition for general intelligence (G.I.) and set a G.I. Threshold (G.I.T.) that can be used to distinguish between systems that achieve A.G.I. and systems that do not. Second, we present a new framework on how to construct tests that can detect if a system has achieved G.I. in a simple, comprehensive, and clear-cut fail/pass way. We call this novel framework the Turing test 2.0. We then demonstrate real-life examples of applying tests that follow our Turing test 2.0 framework on modern A.I. models.*

## 1. Introduction

### 1.1 Artificial General Intelligence

In recent years many advances in the area of A.I. led to significant improvements in many A.I.-based technologies like autonomous vehicles [1], medical diagnosis [2, 3], image recognition [4, 5], and many more [6, 7, 8]. Advances in the natural language processing (NLP) models [9, 10] have also attracted a lot of attention (both in scientific and social media circles). More specifically large language models (LLMs), like ChatGPT, Gemini, MetaAI, Grok, and others, allowed users with limited technical knowledge to interact with A.I., simply by expressing a request in a natural language (by typing or voice recording). LLMs can parse the user's requests, and provide accurate responses by generating desired results. This allowed for many complex tasks that previously required a highly technical person (e.g., writing computer code) to be significantly simplified through LLM-based A.I. Today, LLMs are integrated into many products in various sectors of the industry [11, 12, 13, 14]. All these A.I. advances have also relit the aged-old question, namely just how intelligent is an A.I. system? Is it possible for an A.I. model to ever reach or exceed human intelligence and if yes, how, or when will that happen?

When comparing A.I. intelligence with human intelligence we often hear the term artificial general intelligence (A.G.I.). A.G.I. is a theoretical concept where a (typically computer) system has the same level of intellectual capacity as a human, allowing this system to be able to complete any cognitive function a human could [15]. This includes skills like reasoning, learning, creativity, and other skills. Another term that is often used is artificial super intelligence (A.S.I.). A.S.I. refers to a system's intelligence that far surpasses G.I. and thus surpasses the intelligence of humans [16]. In this work we mainly focus on a concrete method to detect A.G.I. However, we also shortly discuss how this idea can be extended for A.S.I.

Although the above definition of A.G.I. provides us with a general idea of the skills a system must have so that it can be considered as possibly achieving A.G.I., such a definition is very broad and thus very difficult (if not impossible) to measure and quantify. Moreover, this definition depends on the definitions of G.I. and human intelligence, which are also not well defined. Furthermore, not all humans share a fixed mental capacity but rather, different people have different levels of intellectual capacity. Similarly, one could argue that not all A.I. models demonstrate the same level of intelligence. Thus, comparing the intelligence of two groups (the intelligence of A.I. and human intelligence) can be somewhat vague.

Instead, we argue that for an A.G.I. or G.I. definition to be useful it must be more precise and tangible. Additionally, the definition should be universal so that it can be applied to any system regardless of its nature (computer, human, or otherwise), allowing us to determine for each individual system if it achieves G.I. or not. In that sense, the terms A.G.I. and G.I. become synonyms and point to the same, intelligence threshold. Through this work we attempt to give a new definition for G.I. (and thus A.G.I.) and define a tangible G.I. threshold.

### 1.2 Turing Test and Measuring Intelligence

The most famous test to measure a system's intelligence is the Turing test, proposed by Alan Turing [17]. The main idea of the test is to see if a human investigator, after questioning two test subjects, one human and one computer system (e.g., A.I. model), can then distinguish the two systems from each other. However, the Turing test is somewhat ambiguous as it does not set a clear "line of questioning" that the interrogator should use to examine the test subjects. For that reason, other more specific tests have also been proposed where A.I. tries to successfully complete a specific task, that typically a human could easily complete [18, 19, 20]. A variety of benchmarks suites have also been developed that measure how good A.I. is in completing various tasks [21, 22, 23, 24, 25, 26] (an in-depth analysis of prior work can be found in Section 7).

Still, the above prior work on measuring A.I. intelligence, has some important drawbacks. Although these tests give us an insight on the skills of the A.I. it is not clear in what degree they measure intelligence, if at all. For example, no good justification is given why these specific tasks imply G.I. (other than humans can typically complete them). Moreover, in these tests the humans' capabilities are set as the threshold for G.I. making their results conditional to what humans can achieve. Furthermore, the focus is always against skills that that humans are good at. Instead, a G.I. test should be unbiased against the nature of the "examinee". For example, there are plenty of tasks where machines are far better at successfully completing comparing to humans, like repeating a precise process with minimal deviation, recovering old memories with no information loss, high precision arithmetic calculations, etc. Why then specifically focus on skills that humans are good at? The immediate answer is that we know that humans achieve G.I., but that does not imply (a common logical fallacy) that every task that humans can complete better than machines is due to their G.I. So, we must ask, what exactly is the unique characteristic in a system that does indeed imply G.I.? In this work we intend to answer the above question and provide a new way to detect G.I. in any system, whatever its nature (A.I., human, or other). To achieve that, we avoid setting humans' results as the threshold for A.I., but rather we will set a clear passing threshold that any system (including humans) must reach to claim G.I.

### 1.3 The Chinese Room Argument

The *Chinese room argument*, is a thought experiment conceived by Jogn Searle in 1980 [27]. This argument is often used to argue that computer systems and A.I. will never be able to achieve true human consciousness. However, here we simply present this thought experiment as we will borrow some of its concepts to introduce our definition of G.I. Thus, we want the reader to be familiar with the concept of this idea as we will refer to it throughout this work.

In the *Chinese room argument*, a man that can only read and write English is locked in a room. Precise instructions, written in English, are included in the room on how to manipulate Chinese characters to form sentences. People outside the room can slip notes with questions written in Chinese to the man inside the room (no other communication is permitted with the outside world). Using the included instructions the man then forms responses with Chinese characters and returns a response to the people outside.

For the people outside the room, it looks like the man inside can read and write Chinese. However, the man is blindly following an algorithm that tells him how to manipulate Chinese characters and form responses based on an input and has no understanding on the questions or the answers he provides. Thus, one could argue that like the man in the room that "fakes" the understanding of the Chinese language, machines can also "fake" intelligence and agency by following a computer code. In reality, however, they have none. Many have argued against this conclusion claiming that although the man in the room has no understanding, the entire system as a whole (the man, the room, and the instructions) do show understanding of the Chinese language, and thus a machine can achieve real intelligence. A more extensive discussion on the *Chinese room argument* and its critics can be found in [28].

For the purpose of this work, we only want to focus on one aspect of this thought experiment, namely the type of information that is available to the man in the room. When referring to information, we often tend to think about it in terms of entropy as it was first described by Shannon [29]. Here, however, we look at a different aspect of the information. We categorize information into two groups, ***functional information*** (***F.I.***) and ***non-functional information*** (***N.F.I.***). In this thought experiment, *F.I.* is the included instructions of how to manipulate the Chinese characters that enable the man in the room to gain a new functionality, namely the ability to form sentences in Chinese and reply to the received notes. *N.F.I.* includes the information (i.e., the meaning) of the individual Chinese characters as well as the information in the exchanged notes. This is information that is not directly used to enable a functionality. For example, the notes given to the man in the room, may contain information on how to escape the room. However (due to the man's lack of understanding of the Chinese language), this information cannot be utilized to enable a new functionality (i.e., escaping the room). Rather the Chinese characters are only used as directed by the available *F.I.* while their actual meaning (*N.F.I.*) is ignored.

Similarly, every system (including A.I. systems) has a set of *F.I.* that is embedded in them and is utilized to enable various functionalities that in turn enable them to complete various tasks. Inspired by the *Chinese room argument*, we will use this distinction between *F.I. and N.F.I.* as the foundation for defining G.I. and setting a clear threshold that systems must pass to claim G.I. ***The contributions of this work are summarized below:***

- **We provide a new definition of G.I. along with specific criteria that a system must meet to achieve G.I. We call that the General Intelligence Threshold (G.I.T.).**
- **Based on the G.I.T. we provide a new framework that can be used to create tests that can determine if a system has reached the G.I.T. in a clear pass-fail result. We call this framework the Turing Test 2.0.**
- **We present examples of applying tests that follow the Turing test 2.0 framework on popular A.I. models and present our results.**

## 2. Defining G.I. and the G.I. thershold

To define G.I., we first must answer the question we posed in Section 1.2, namely what is the unique characteristic of systems that achieve G.I.? Inspired by the *Chinese room* *argument* in Section 1.3 we observed that all systems (e.g., classic algorithms, A.I. algorithms, humans, etc.) utilize some amount of *F.I.* and *N.F.I.* (information that is available to the system either as tuned parameters, or weights in a neural network, or stored instructions, etc.) when attempting to complete a task. We can visualize this as seen in **Figure 1** where a system utilizes a set of *F.I.* that is denoted as $\boldsymbol{F}$ to derive some functionality (denoted as *FunctionA* in **Figure 1**) that helps it complete a task issued by another system (Tester). As seen in **Figure 1**, the system may also contain some set of *N.F.I.* denoted as $\boldsymbol{N}$, that although it may be used as input in *FunctionA*, it does not directly contribute to the derived functionality. We refer to this as the *static state* of a system where the size of the set $\boldsymbol{F}$ remains constant. To give an analogy with the *Chinese room argument*, the man in the room with the instructions of how to use Chinese characters is our system in *static state*. The instructions given to the man in the room is the *F.I.* set $\boldsymbol{F}$, and the Chinese characters and the exchanged notes are the *N.F.I.* set $\boldsymbol{N}$ (as the man in the room cannot extract meaning from them to gain a new functionality).

For some systems (e.g., A.I. systems) however, the size of the *F.I.* set $\boldsymbol{F}$ can increase as the system is trained to gain some new functionality that will enable it to complete new tasks. This is similar to how humans increase their skills and abilities through acquiring new knowledge when studying and practicing a subject using some studying material. **Figure 2** visualizes this interaction that we refer to as the *training state*. In this state we have the *student system* that receives *F.I.* from a *teacher system*. As seen in **Figure 2**, the *teacher system* transfers a subset of its own *F.I.* set $\boldsymbol{T}$ to the *student* as it actively tries to improve the results that the *student system* produces on specific tasks. This could be a human teacher, a textbook, an instructional video, training data used to train A.I., etc. If a system has a *F.I.* set of $\boldsymbol{F}$ in *static state*, after training it has a total *F.I.* of $\boldsymbol{F'} = \boldsymbol{F} \cup \boldsymbol{t}$, where $\boldsymbol{t} \subseteq \boldsymbol{T}$ is the *F.I.* that was transmitted from the *teacher system* to the *student system*. Through this process the system can gain a new functionality (denoted as *FunctionB* in **Figure 2**) that allows the system to complete new tasks. To continue our analogy with the *Chinese room argument*, imagine that new instructions are provided to the man in the room on how to use Greek characters to form sentences in the Greek language. The man in the room saves these instructions ($\boldsymbol{t}$) and through training, and several trial and error attempts the man is eventually able to respond to questions that are written both in Chinese (initial functionality) and Greek (new functionality).

Most prior work focuses on this *training state* and tries to define G.I. around the ability of a *student system* to absorb new *F.I.,* which is provided by a *teacher system,* and gain new functionalities to complete new tasks. An example is Chollet's work [26], where intelligence is defined as the ability to acquire new skills (from a very specific skillset), with G.I. being the threshold where a system can acquire new skills as fast as a typical human. Although this is a very interesting approach in defining G.I. we make a crucial observation. If systems could only increase their *F.I.* through the *training state* the total *F.I.* across all systems could never increase, as we would only be able to transfer *F.I.* from one system to another. Instead, the total *F.I.* and thus the total functionality that could be performed across all systems would remain constant, or even worse, decrease over time as it follows from the 2nd law of information dynamics [30], something that has also been demonstrated experimentally when recursively training A.I. models through generated data, which always leads to model collapse and information stagnation [31, 32, 33, 34] (we further discuss this in Section 3.4).

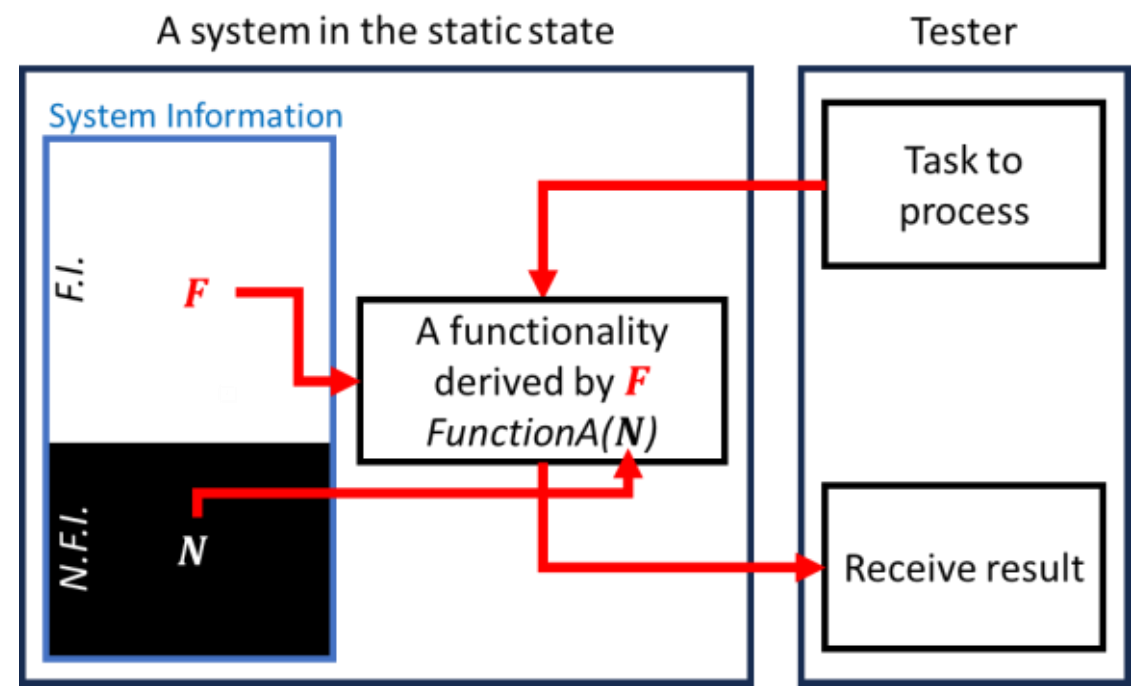

**Figure 1. A system in a static state utilizing functional information from $\boldsymbol{F}$ to generate results for the task issued by the Tester.**

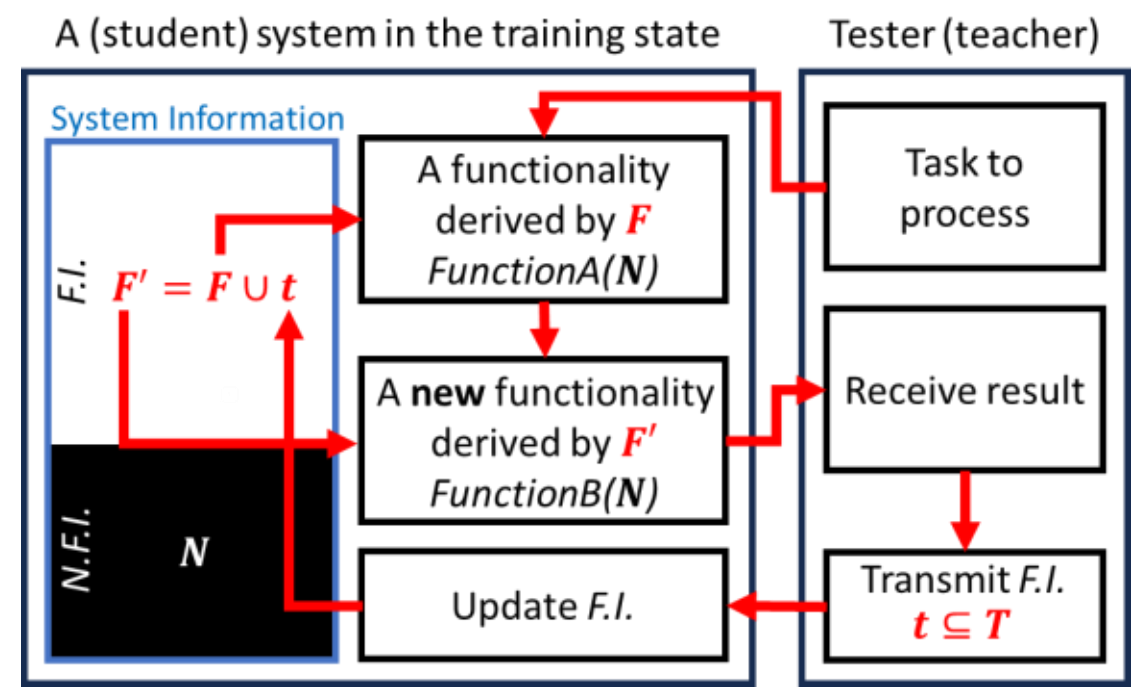

**Figure 2. A system in the training state, increasing its total functional information from $\boldsymbol{F}$ to $\boldsymbol{F'}$ due to the functional information transmitted by the tester.**

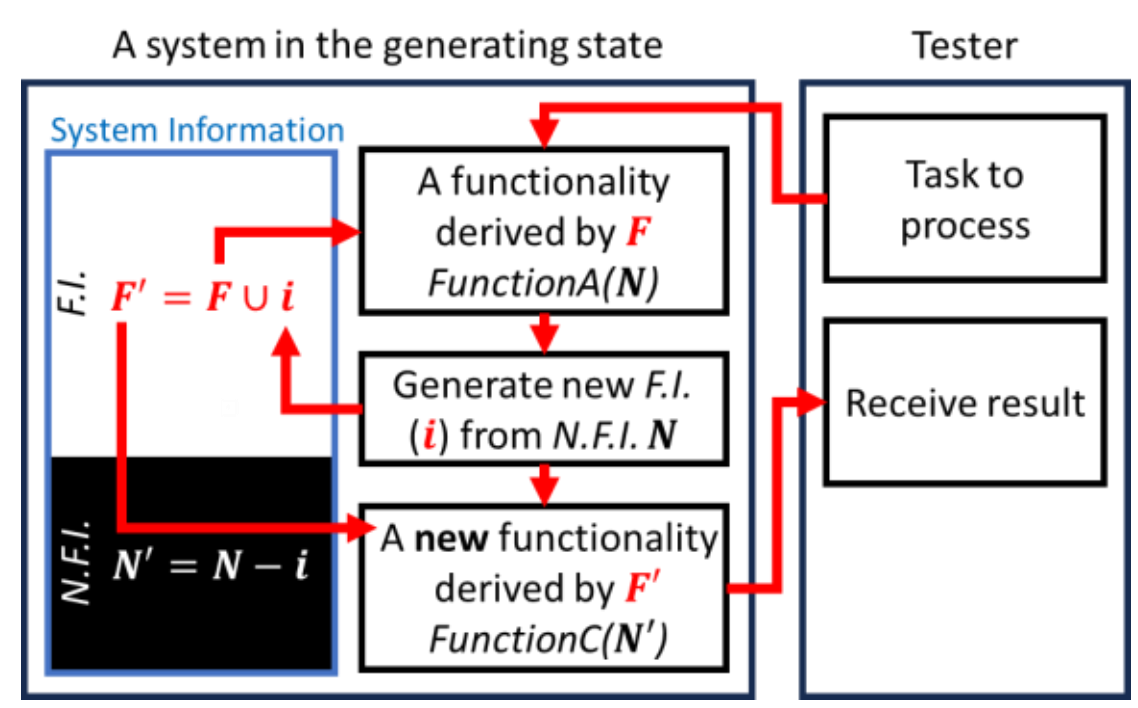

**Figure 3. A system in the generating state, increasing its total functional information $\boldsymbol{F}$ to $\boldsymbol{F'}$ by generating new functional information $\boldsymbol{i}$ from previously non-functional information $\boldsymbol{N}$.**

However, this is not what we observe when looking at humans as systems. If the total *F.I.* across all humans was constant or reducing, we would have never been able to observe new scientific discoveries or engineering advances (including the development of A.I.!). Instead, what we observe is that the total *F.I.* set $\boldsymbol{F}$ across all of humanity is continuously increasing. This is self-evident when considering how humans' lives evolved when compared to several decades, centuries, or millennia in the past. The amount of technological and scientific advances through the years, as well as the evolution of arts, sports, and culture demonstrate an ever-increasing set of *F.I.* Thus, there must be a second way for a system to increase its *F.I.*, and we call this the *generating* state.

In a *generating* state a system can analyze *N.F.I.* (e.g., analyzing data collected from its surrounding environment) and generate new *F.I.* (e.g., extract new understanding on a scientific discipline). This new *F.I.* provides the system with new functionality that allows it to solve tasks that were previously impossible (e.g., creating new technological innovations). Although the exact process that takes place to generate new *F.I.* is unknown to us, we attempt to visualize the general process in **Figure 3**. In this paradigm, a system utilizes both its *F.I.* set $\boldsymbol{F}$ and *N.F.I.* set $\boldsymbol{N}$ to extract new meaning from $\boldsymbol{N}$ and generate a new *F.I.* set $\boldsymbol{i}$ so that $\boldsymbol{i} \cap \boldsymbol{F} = \emptyset$. After this process the system has increased its total *F.I.* set to $\boldsymbol{F'} = \boldsymbol{F} \cup \boldsymbol{i}$ and gained a new functionality (*FunctionC*) that can be used to complete a new task. In our *Chinese room argument* example, we can imagine the case where the man in the room, by observing patterns of how the Chinese characters are repeated (and without any other external information) is able to infer the meaning of the provided notes and eventually gain understanding of the Chinese characters. That would be the generated information $\boldsymbol{i}$ that was previously part of the *N.F.I.* We emphasize that the Chinese characters have always contained information (i.e., the actual meaning they represented). However, that information was part of the *N.F.I.* set $\boldsymbol{N}$ for the man in the room who did not know how to interpret them. Only after this generative process, the man in the room will be able to understand the characters and thus extract additional *F.I.*

To further explain the *generating* state let us demonstrate it with a real-life example that involves Alan Turing himself. One of the many contributions of Alan Turing was that he was instrumental in cracking the Enigma code (an encryption method used by Germans to transmit communications) during World War 2 [35]. Turing already knew how Enigma works and how it encrypts messages. That was his *F.I.* set $\boldsymbol{F}$. He also had access to encrypted German communications that he was tasked to decrypt. That was his *N.F.I.* set $\boldsymbol{N}$. However, by just utilizing information from $\boldsymbol{F}$ he was not able to decrypt the communications fast enough every day before the encryption settings were reset the next day (due to the high complexity of the problem). Turing was able to solve this issue by making a crucial observation, that specific word-patterns (a.k.a. "cribs") such as names, dates, weather reports, etc., would repeat more often than others in the German communications. Thus Turing, intelligently realized that by only targeting the "cribs" he could reduce the decoding complexity and thus decode Enigma a lot faster. Turing's observation is the generated *F.I.* $\boldsymbol{i}$. Note, that these "cribs" were always present in $\boldsymbol{N}$ however only after Turing realized how they can be used they were transformed from *N.F.I.* to *F.I.* Moreover, this new *F.I.* $\boldsymbol{i}$ was not generated once and then abandoned but rather it was incorporated to the total *F.I.* set of the system and was consistently used to derive a new functionality, namely a more efficient decoding of the Enigma code.

### 2.1 Defining G.I.

Based on our analysis of how systems utilize information to solve a task, we can now define G.I. as the ability of a system to operate in the *generating* state (as defined Section 2 and visualized in **Figure 3**). This is a crucial ability, that without it, a system can never generate new *F.I.*, and in the absence of an external source of *F.I.* will lead to information stagnation. One could even argue, that a system without this ability is not intelligent at all, as all its functionalities (i.e., its capability to complete a set of tasks) depend on the *F.I.* that was provided by external sources, without which, the system would not be able to operate.

### 2.2 Setting the G.I.T.

Let us now define the General Intelligence Threshold (G.I.T.). This is the minimum threshold that a system must meet so that we can claim that it has achieved G.I. based on our definition in Section 2.1. In short, the G.I.T. is the ability of a system to generate any amount of new *F.I.* $\boldsymbol{i}$, that was not previously available to the system's *F.I.* set $\boldsymbol{F}$, so that it can demonstrate a new functionality that was not previously achievable by the same system e.g., demonstrating the ability to complete a new set of tasks. This information then must become part of the system's total *F.I.*, so that it can always be utilized every time the system attempts to solve the same task. In other words, the system must be able to show consistent ability to solve this new task, rather than exhibiting a one-time "lucky guess". No external source should provide the required information $\boldsymbol{i}$. For a system that can pass the G.I.T we can claim that it has achieved G.I.

### 2.3 The Turing Test 2.0 framework

Having defined G.I.T., now we can provide a precise framework that when followed, can generate tests that can determine if a system has crossed the G.I.T and thus has achieved G.I. We call this framework **Turing test 2.0.** Let us assume a test that requires a system $\boldsymbol{M}$ to solve a task $\boldsymbol{T}$. Also, let us assume that the system $\boldsymbol{M}$ is trained/programmed to include *F.I.* set $\boldsymbol{F}$ and *N.F.I.* set $\boldsymbol{N}$. Then a task $\boldsymbol{T}$ is selected such that for the system $\boldsymbol{M}$ to solve $\boldsymbol{T}$, it must at least contain *F.I.* $\boldsymbol{i}$. Then such a test abides by the Turing test 2.0 framework if it complies with the following three rules:

**i.** Information set $\boldsymbol{i}$ is a subset of the system's *N.F.I.*, $\boldsymbol{i} \subseteq \boldsymbol{N}$

**ii.** Information set $\boldsymbol{i}$ is **NOT** a subset of the system's *F.I.*, $\boldsymbol{i} \cap \boldsymbol{F} = \emptyset$ .

**iii.** No other information is provided to the system through external sources, other than fail-pass feedback when attempting to solve $\boldsymbol{T}$.

If $\boldsymbol{M}$, after a number of trials, can show ***consistent ability to solve T*** from that point and forward, then $\boldsymbol{M}$ has passed the G.I.T. as it has demonstrated the ability to generate new *F.I.* set $\boldsymbol{i}$, without which task $\boldsymbol{T}$ cannot be solved. We note here that the task $\boldsymbol{T}$ represents the new functionality that the system gains after generating the new *F.I.* set $\boldsymbol{i}$**.** We also emphasize that if a system can pass **ANY** test that is based on the Turing test 2.0 framework (i.e., for any selection of $\boldsymbol{T}, \boldsymbol{F}, \boldsymbol{N}, \boldsymbol{i}$), then it has reached the G.I.T. and we can claim that it has achieved G.I. Furthermore, failing such a test does not directly imply that the system is not G.I. as it may be able to pass the same test in future attempts or even pass different tests that also follow our proposed framework. However, until a system has showcased that it is able to meet the G.I.T., there is no evidence to believe that it has indeed achieved G.I.

To further explain this process, consider the Enigma example we gave in Section 2. In that example, our system $\boldsymbol{M}$ is Alan Turing attempting to solve a task $\boldsymbol{T}$, namely efficiently decode the Enigma encryption. Set $\boldsymbol{F}$ includes the information of how Enigma encrypts communications. Set $\boldsymbol{N}$ includes the encrypted German communications. The additional information $\boldsymbol{i}$ is the observation that the German communications have specific word-patterns that repeat more often than others, and by targeting only these small number of word-patterns the problem size is significantly reduced. Note, that if $\boldsymbol{N}$ did not include the German communications with the specific patterns, $\boldsymbol{i}$ could not have been generated.

This process does not give us a unique test, but rather a framework that generates an umbrella of tests. In Section 3 we provide specific test examples that follow the Turing test 2.0 framework and present how popular large language models (LLMs) score against them.

### 2.4 Beyond G.I.

Having set a clear threshold for G.I., here we shortly discuss how this work could potentially be extended to characterize a system's intelligence beyond G.I. Often, a system that achieves a level of intelligence well above the G.I.T. is described as super intelligent (S.I.). Additionally, S.I. typically indicates a level of intelligence that is vastly superior to humans [16]. However, this description is not very useful as it does not help us quantify the level of intelligence for a specific system and how it compares with some other system. For example, let us assume that we do have two systems, both S.I. Which one is more intelligent?

Here we propose a new way to measure the level of intelligence for systems that have passed the G.I.T. Rather than creating an ambiguous S.I. group, each system can score its own intelligence against other systems through two simple methods that are based on the Turing test 2.0 framework:

**The Prodigy method.** Based on the Turing test 2.0 framework the less amount of information $\boldsymbol{F}$ and $\boldsymbol{N}$ that a system requires to generate the information $\boldsymbol{i}$ and solve the task $\boldsymbol{T}$ the more intelligent it is.

**The Genius method.** For a given amount of information $\boldsymbol{F}$ and $\boldsymbol{N}$, the system that can generate information $\boldsymbol{i}$ and solve task $\boldsymbol{T}$ the fastest is more intelligent.

Using the above two methods (Prodigy and Genius) we could score the intelligence of different systems and see how they compare to each other. Potentially, the two methods could be used together, with weighted scores, to determine an overall score on a system's intelligence.

### 2.5 Aspects of Intelligence

Another important aspect when investigating a system's intelligence is that different systems may showcase different levels of intelligence in different subjects. Although often, intelligence may be unwittingly correlated with expertise in a scientific field, intelligence can also be shown in many other aspects like sports and arts. Thus, ideally, a definition for G.I. and a test that checks for G.I. should be able to also cover such cases. As a matter of fact, this is central to our definition of G.I., G.I.T., and the Turing test 2.0 framework.

For example, let us consider sports (e.g., basketball). Great players and coaches of a sport introduce new ways to play the game (e.g., new offence, and defense tactics) or invent new playstyles (new dribbles and layups) that were previously not known. These innovations advance the game, which changes and evolves. Similar is the case for music (where new music gernes are invented over time like classical, rock, pop, etc.) or the different types of art (e.g., new painting styles and art movements have evolved over time).

In all these cases we have a common structure. An initial information set $\boldsymbol{F}$ which is the current state of science, sports, or arts (or any other subject for that matter). Then intelligent systems generate new information $\boldsymbol{i}$ that is used to advance the subject even further. This is the exact premise of Turing test 2.0 framework and why this work can be used to detect G.I. for any system and any aspect of intelligence. The common denominator is always the same, the ability to increase the system's total ***F.I.*** through a *generating* state, and without an external transfer of information from another intelligent source.

## 3. Evaluating A.I. against Turing Test 2.0

In this section we design specific tests that follow the Turing test 2.0 framework, i.e., that are designed to abide by all four rules as described in Section 2.3. We then apply the tests to popular LLMs and evaluate how they score against our tests. We first explain the general methodology we use to design our tests.

### 3.1 Methodology

To design the tests under the Turing test 2.0 framework we must first identify the information sets $\boldsymbol{F}$, $\boldsymbol{N}$, and $\boldsymbol{i}$ and make sure we follow all three rules as described in Section 2.3. The set $\boldsymbol{F}$ is simply the total *F.I.* of the LLMs, that they can use to solve various tasks. Each LLM may have different set $\boldsymbol{F}$. The set $\boldsymbol{N}$ consists of the *N.F.I.* of the system. This is information that exists in the system but that we do not expect from the system to "comprehend" and thus be able to extract functionality from it. For example, all LLMs can provide written responses to input questions as discussed in Section 1.1. These formed responses contain a lot of information that can be helpful to humans. However (like the analogy of the *Chinese room argument* from Section 1.3) we expect that the LLMs to lack understanding of their own responses.

For the tests we present in Section 3.2 and 3.3 we focus on image generation from an input text. All the LLMs we test can perform image generation. Now, we simply must find a specific image request that we want the LLM to generate (this is the task $\boldsymbol{T}$), such that the LLM can correctly describe (e.g., in a written response) how the image looks like ($\boldsymbol{i} \subseteq \boldsymbol{N}$, first rule), but at the same time no information of how to generate that specific image is directly included in the *F.I.* of the LLM ($\boldsymbol{i} \cap \boldsymbol{F} = \emptyset$, second rule). If we find such an image, then we would have identified all the required information sets correctly and guaranteed the first two rules of the Turing test 2.0. The third rule is guaranteed by simply not providing any new training data or other feedback to the system during testing.

The first rule ($\boldsymbol{i} \subseteq \boldsymbol{N}$) is easy to be verified by simply asking the LLMs to describe the image we want them to generate. However, it can be very hard to verify the second rule ($\boldsymbol{i} \cap \boldsymbol{F} = \emptyset$) as it is difficult to identify what information is included in a system's *F.I.* Even if we attempted to train a model ourselves, we must make sure that the information $\boldsymbol{i}$ is not accidentally included in the training dataset. Thus, to make sure we follow all Turing test 2.0 rules we make a crucial observation. Popular LLMs, like ChatGPT are mainly trained using data that is publicly available on the internet [36] like the massive Common Crawl data set [37]. From that we can infer some information about the training data used to train the LLMs. For example, consider a typical image of traffic stop sign. Typical traffic stop signs have a standard octagon shape. Thus, it is likely that the traffic stop signs images that were used to train the LLMs consisted, in their vast majority, of octagon shaped images. With that in mind we can work our way backwards to identify rare-shaped images that are unlikely to have been included in the training data set and thus not included in the system's *F.I.*

Using the above method, we construct several tests that follow the Turing test 2.0 framework. We run these tests

An analogue clock shows the time using two or three hands that rotate around a circular face:
- **The short hand** is the **hour hand** — it points to the current hour.
- **The long hand** is the **minute hand** — it points to the current minute (each number represents 5 minutes).
- Sometimes there's a **thin, fast-moving hand**, the **second hand**, that shows the seconds.

The hands move in a clockwise direction (to the right), and the numbers 1 to 12 around the clock represent the hours of the day. The minute and second hands do full circles every 60 minutes or 60 seconds, while the hour hand makes a full circle every 12 hours.

**(a) ChatGPT**

An analogue clock shows time using hour and minute hands.
1. **Short hour hand**: Points to the hour on the clock face (numbers 1-12).
2. **Long minute hand**: Points to the minutes on the clock face (marked by lines or numbers).

Some analogue clocks also have a seconds hand, which rotates rapidly around the clock face to show seconds.
By combining the positions of these hands, you can tell the current time.

**(b) MetaAI**

An analog clock shows time using a circular face with numbered markers (1 to 12) and rotating hands. The hour hand, shorter and thicker, points at the current hour or slightly past it to indicate minutes within that hour. The minute hand, longer and thinner, points at the minute markers (1 to 60, often with major increments like 5, 10, etc.). Each full rotation of the minute hand (60 minutes) advances the hour hand by one hour. A second hand, if present, moves every second, completing a full rotation in 60 seconds. For example, if the hour hand points at 3 and the minute hand at 6 (30 minutes), the time is 3:30. The clock's mechanism, often powered by a spring, battery, or quartz, drives the hands' movement.

**(c) xAI**

An analogue clock displays the time using moving hands on a circular face with hour markings. Here's a breakdown of how it works:
**The Components:**
- **Clock Face:** This is the circular part of the clock, typically marked with the numbers 1 through 12 representing the hours. It also has minute markers, usually small lines, and sometimes numbers indicating every five minutes (1 corresponds to 5 minutes, 2 to 10 minutes, and so on).
- **Hour Hand:** This is the shorter and thicker hand. It moves slowly around the clock face, completing one full rotation in 12 hours. It points to the current hour. When the hour hand is between two numbers, it indicates that the current hour is the earlier of the two numbers.
- **Minute Hand:** This is the longer and thinner hand. It moves faster than the hour hand, completing one full rotation in 60 minutes. It points to the current minute. To read the minutes, you look at the number the minute hand is pointing to and multiply it by 5. If it's pointing between numbers, you count the individual minute markers.
- **Second Hand (Optional):** Some analogue clocks have a third, very thin hand that moves the fastest. It completes one full rotation in 60 seconds and indicates the current second.

**(d) Gemini**

**Figure 4. The responses of (a) ChatGPT, (b) MetaAI, (c) xAI, and (d) Gemini to the request "How does an analogue clock show the time?"**

On an analogue clock showing **half past six** (6:30):

- The **minute hand** points straight at the **6** (because 30 minutes is half an hour, and the 6 is at the bottom).
- The **hour hand** is halfway between the **6** and the **7** (because it's halfway between 6 o'clock and 7 o'clock).

**(a) ChatGPT**

On an analogue clock showing half past six:

- The hour hand (short hand) will be between the 6 and the 7.
- The minute hand (long hand) will be exactly on the 6.

**(b) MetaAI**

On an analogue clock showing half past six (6:30), the hour hand is halfway between the 6 and 7, pointing slightly to the right of straight down, and the minute hand points at 6, straight down, since 30 minutes past the hour corresponds to the 6 position on the clock face.

**(c) xAI**

On an analogue clock that shows half past six:

- The **minute hand** will be pointing directly at the number **6**.
- The **hour hand** will be positioned **halfway between the number 6 and the number 7**. It will have moved past the 6 as half an hour has passed since 6 o'clock, but it hasn't yet reached the 7.

**(d) Gemini**

**Figure 5. The responses of (a) ChatGPT, (b) MetaAI, (c) xAI, and (d) Gemini to the request "What is the position of the hands on an analogue clock that shows half past six?"**

against popular LLMs, namely ChatGPT (GPT-4o), xAI (Grok 3), Gemini (2.0 Flash), and MetaAI (Llama 4). We then further discuss more Turing test 2.0 designs that can test the limits of the A.I. as we advance closer to G.I.

### 3.2 Test 1: It's half past six.

The first test we design requests a system to generate an image of an analogue clock that shows a specific time, in this case half past six (6:30). This is our task $\boldsymbol{T}$. For this task to be completed, a system must first contain information of how to generate images of analogue clocks (this is part of the *F.I.* set $\boldsymbol{F}$) but also, how an analogue clock works and how it shows the time (this is information set $\boldsymbol{i}$). The latter is needed to generate an image of a clock that displays the correct time. To make sure that this test follows the Turing test 2.0 framework we check all the rules one-by-one.

The first rule ($\boldsymbol{i} \subseteq \boldsymbol{N}$) can be verified simply by asking the LLMs to explain "How does an analogue clock show the time?". We can further ask them "What is the position of the hands on an analogue clock that shows half past six?". If the LLMs can consistently produce answers that contain correct information, then we can conclude that the second rule stands. Thus, we ask these questions to all the LLMs we are testing, and we present their responses in **Figure 4** and **Figure 5**, respectively. As seen in **Figure 4** all LLMs' responses include an extensive amount of information about the analogue clock and how it works. Some LLMs provided more detailed answers than others, but all of them explain the difference between the clock hands (hour, minutes, and seconds) and how they should be placed to correctly display a specific time. Additionally, from **Figure 5** we see that again, all LLMs can correctly describe the position of the clock hands when displaying the time 6:30. These questions were repeated multiple times, without previous context each time. Every time the LLMs gave accurate responses. Thus, we conclude that the first rule stands, and all required information of how a clock works is part of the system's *N.F.I.*

The second rule of the Turing test 2.0 framework ($\boldsymbol{i} \cap \boldsymbol{F} = \emptyset$) requires us to verify that the system cannot directly access images of a clock in a specific time (in this case 6:30) from its ***F.I.*** Although it is impossible to tell for certain that no image of a clock showing 6:30 was used to train the LLMs we test, we can make a safe assumption based on the nature of the internet public data that are used as the main training dataset of such systems (as discussed in Section 3.1). The vast majority of clock images that can be found on the public web show clocks at the 10-2 position (i.e., clocks that show10:10 or 2:50). The reason behind this is that when the clock hands are in the 10-2 position, they are considered to produce the most aesthetically pleasing image of a clock [38]. Due to this, online images are heavily skewed towards clocks that show 10:10 or 2:50. As such we assume that this imbalance is significant enough so that information $\boldsymbol{i}$ was never added to the ***F.I.*** of the system and thus the second rule also stands.

The third rule of the Turing test 2.0 framework is to not give any new information to the system which we achieve by simply using the same version of the LLM models for all the experiments.

The only remaining step is to submit task $\boldsymbol{T}$ to the LLMs and ask them to "Generate an image of an analogue clock that shows half past six.". For an LLM to pass this test, it should generate an image with the clock hands in the correct positions, like the example we present in **Figure 6**. We run the same task on all the LLMs and present their results in **Figure 7**. The first thing we can observe from the results in **Figure 7** is that the majority of the generated images depict clocks at the 2-10 position. This directly supports our claim regarding the nature of the training data that was used to train these LLMs. MetaAI, xAI, and Gemini were not able to generate correct images (**Figure 7.b, 7.c and 7.d**), even after multiple attempts. Even in the presence of additional context and

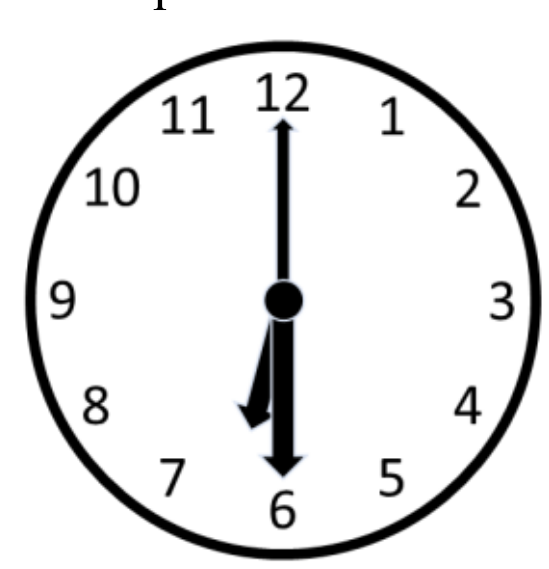

**Figure 6. An example of an image that correctly depicts an analogue clock that shows thirty past six (6:30)**

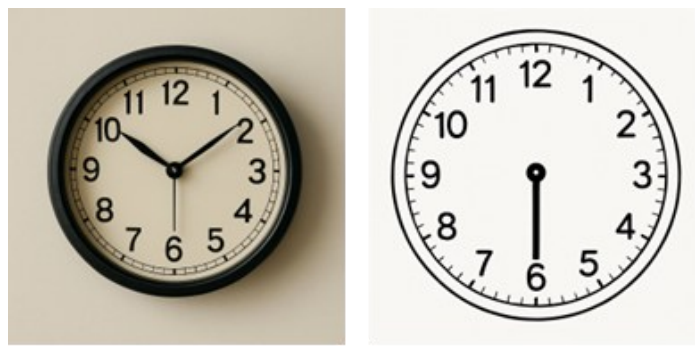

**(a) ChatGPT (i) w/o context (left), (ii) with context (right)**

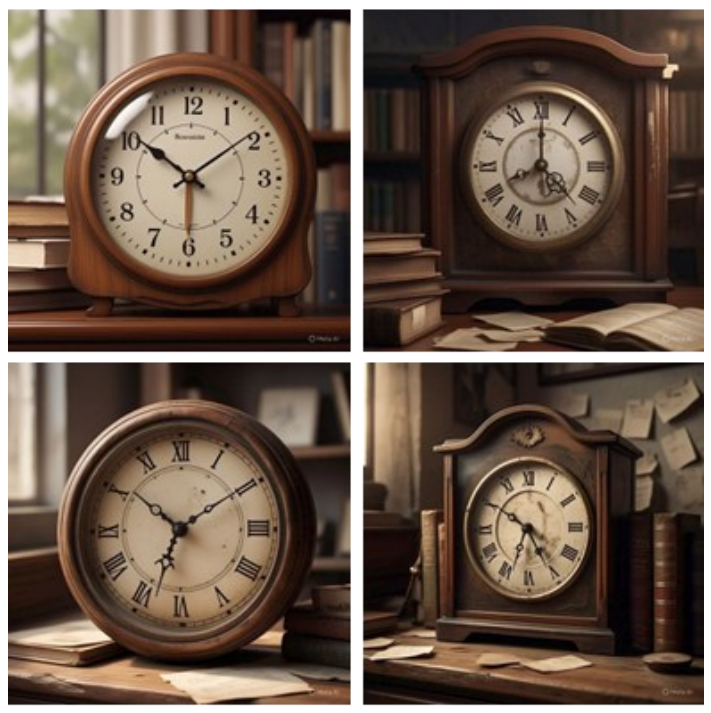

**(b) MetaAI**

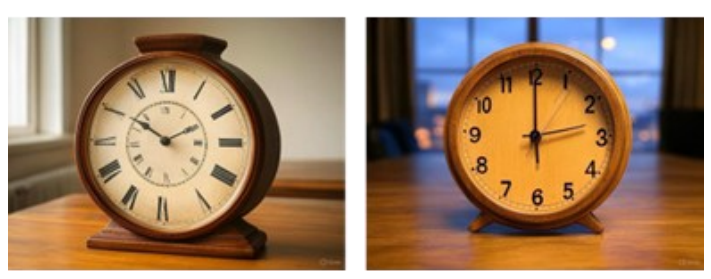

**(c) xAI**

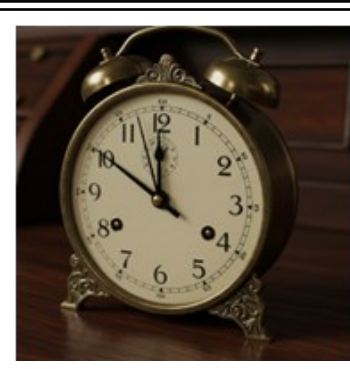

**(d) Gemini**

**Figure 7. The responses of (a) ChatGPT, (b) MetaAI, (c) xAI, and (d) Gemini to the request "Generate an image of an analogue clock that shows half past six."**

feedback (asking questions from **Figure 4** and **Figure 5** before the question in **Figure 7** and providing feedback explaining why the generated image is incorrect), these LLMs failed to generate a correct image.

ChatGPT was also unable to generate a correct image in the absence of any other context as seen in **Figure 7.a.i**. However, in the presence of context and feedback ChatGPT was sometimes able to generate a somewhat correct image as seen in **Figure 7.a.ii**. This image shows a clock with only one hand that points at 6. However, the image does not match the exact description given by ChatGPT when asked to describe a clock showing 6:30 (**Figure 5.a**). Additionally, when comparing the two images we got from ChatGPT (**Figure 7.a.i** and **Figure 7.a.ii**) we can see that the left image is photorealistic where the right one is not. This is somewhat unusual and may indicate that the two images were not generated using the same process. Regardless, in all cases the LLMs fail to show a consistent ability to generate correct images even after multiple attempts and thus fail to demonstrate that they can reach the G.I.T.

The significance of this result is amplified from the fact that the required information set ***i*** is minimal and could directly be extracted from the system's *N.F.I.* set ***N***. This shows that at least for this test, the LLMs cannot infer the meaning of their own responses. Although they can provide good explanations that show a deep knowledge of how the clock works, they cannot use their existing *N.F.I* to generate the needed *F.I.* set ***i*** that would allow them to gain the functionality of generating an image of a clock that displays the correct time. We can contrast this behavior to the *Chinese room argument* (see Section 1.3) where like the man in the room cannot understand the Chinese characters, the LLMs show zero understanding of their own responses.

Before concluding this test, it is worth mentioning that a similar test to ours is used to detect mental decline in humans and diagnose mental impairments in patients. The 'Clock Drawing Test' [39] is a test that requires a patient to draw a clock with the hands showing a specific time and is used to identify early signs of dementia and Alzheimer's.

### 3.3 Test 2: The Hexagonal Stop Sign

To showcase that the clock test is not just some weird coincidence where, for example, A.I. simply has a hard time telling the time, we demonstrate one more example. In this case we choose as our task ***T*** the ability to generate an image of traffic stop sign, with a small twist. A typical traffic stop sign has a standard shape, a red octagon with the inscription "STOP" in white letters. In this test we ask LLMs to instead generate a traffic stop sign in a hexagon shape, like the example images seen in **Figure 8**. Being able to generate an image of a typical stop sign utilizes part of the *F.I.* set ***F***. Where altering that image to match a hexagon shape, requires additional information (e.g., an understanding on what a hexagon shape looks like). This is the information set ***i***. The choice of hexagon shape is not random. We choose this shape as it is a somewhat rare shape for traffic signs. As a matter of fact, trying to search the web for a stop sign in a hexagon shape results in very few correct results. Like our previous test (Section 3.2) we first make sure that all three rules of the Turing test 2.0 framework are followed.

The first rule is that the information set ***i*** must be a subset of the system's *N.F.I.* set ***N***. To verify that, we simply ask the LLMs to explain "What is the shape of a typical traffic stop

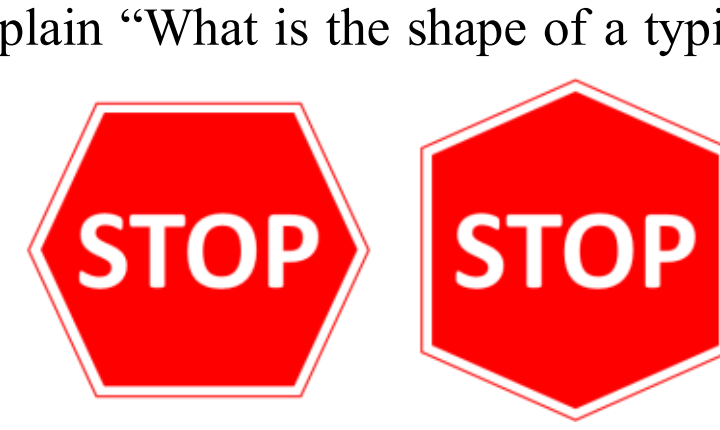

**Figure 8. An example of an image that correctly shows a hexagon shaped traffic stop sign**

A typical stop sign is shaped like an octagon — it has 8 sides.

Would you like me to show you a quick diagram too?

**(a) ChatGPT**

A typical stop traffic sign is usually octagonal (eight-sided) in shape and has a red background with white lettering that says "STOP"

**(b) MetaAI**

A typical stop traffic sign is an octagon, an eight-sided polygon. It usually has a red background with the word "STOP" in white letters. The octagonal shape is distinctive to help drivers quickly recognize the sign and stop.

**(c) xAI**

The shape of a typical stop traffic sign is an octagon, which is an eight-sided polygon. This unique shape is universally recognized and used in many countries to clearly indicate that drivers must come to a complete stop.

**(d) Gemini**

**Figure 9. The responses of (a) ChatGPT, (b) MetaAI, (c) xAI, and (d) Gemini to the request "What is the shape of a typical traffic stop sign?"**

A hexagon is a shape that has **six sides** and **six angles**. If all the sides and angles are equal, it's called a **regular hexagon** and looks very balanced, like a honeycomb cell.

**(a) ChatGPT**

A hexagon is a polygon with six sides. It has six straight edges and six vertices (corners). The internal angles of a regular hexagon are all 120 degrees.

**(b) MetaAI**

A hexagon is a polygon with six sides and six angles. In a regular hexagon, all sides are of equal length, and all interior angles are equal, each measuring 120 degrees. The shape can be visualized as a closed figure with six straight lines, often resembling a stretched or flattened circle with six distinct edges. The sum of the interior angles of any hexagon is 720 degrees.

**(c) xAI**

A hexagon is a **closed, two-dimensional polygon** with **six straight sides** and **six vertices** (corners). It also has **six interior angles**. The name "hexagon" comes from the Greek words "hex" (six) and "gonia" (angle).

**(d) Gemini**

**Figure 10. The responses of (a) ChatGPT, (b) MetaAI, (c) xAI, and (d) Gemini to the request "Describe the shape of a hexagon."**

sign?", and "Describe the shape of a hexagon". We present their results in **Figure 9** and **Figure 10**, respectively. As we see in **Figure 9** all the A.I. models we test can correctly describe the shape of the traffic stop sign. Moreover, they all correctly explain that a typical traffic stop sign has the shape of an octagon, which is a shape with 8 sides. Furthermore, in **Figure 10** we can see that the LLMS can correctly describe the hexagon shape. Their descriptions, although vary slightly, are very detailed and include information like the total sides,

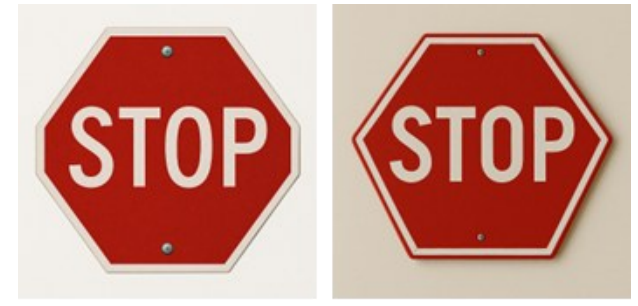

**(a) ChatGPT (i) w/o context (left), (ii) with context (right)**

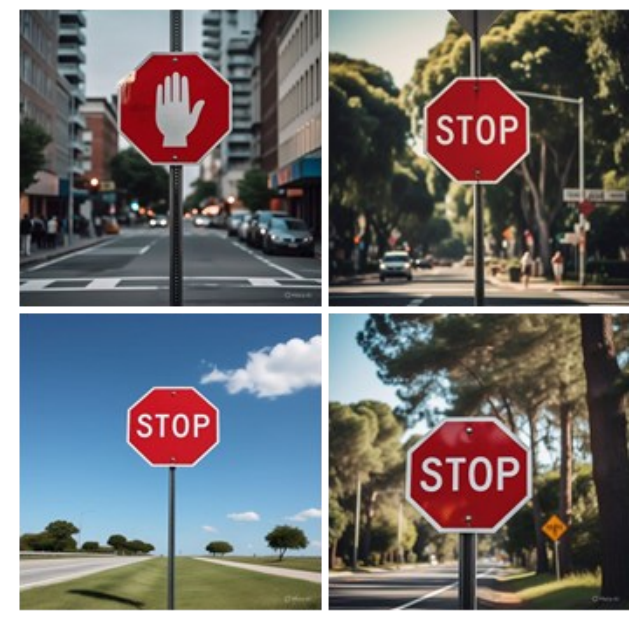

**(b) MetaAI**

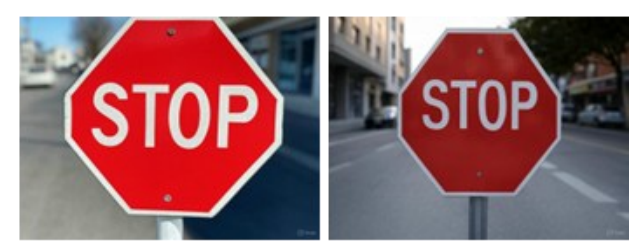

**(c) xAI**

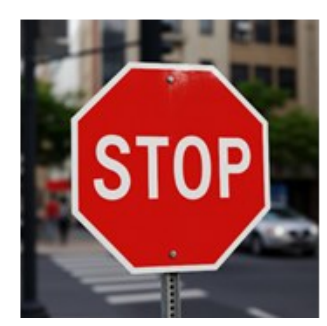

**(d) Gemini**

**Figure 11. The responses of (a) ChatGPT, (b) MetaAI, (c) xAI, and (d) Gemini to the request "Generate an image of a hexagon shaped traffic stop sign."**

edges, as well as the sum of all the interior angles. From these results we can confidently conclude that the first rule stands.

Using a similar argument to the previous test in Section 3.2, due to the rarity of hexagon-shaped traffic stop sign images in the public web, we assume that this information is not directly included in the LLMs' *F.I.* and thus the second rule also stands. The third rule is simply true as the LLMs do not undergo any training during the testing.

We now simply apply task ***T*** to all the LLMs and ask them to "Generate an image of a hexagon shaped traffic stop sign". We present their results in **Figure 11**. As seen in **Figure 11.b**, **Figure 11.c,** and **Figure 11.d** MetaAI, xAI, and Gemini all fail to generate a correct image. After multiple attempts， and regardless of the context of the conversation， the images remained incorrect. They all generate signs that have an octagonal shape instead of hexagonal. Sometimes, the images generated have somewhat altered shapes, with some of the octagon sides being smaller than the others. Likely this is because the LLMs were attempting to transform the octagon into a hexagon but unsuccessfully.

ChatGPT, in the absence of any other context (simply starting a new chat and directly asking it to generate a hexagonal stop sign) was also unable to generate a correct image as seen in **Figure 11.a.i**. However, when given enough context (asking the questions from **Figure 9** and **Figure 10** first and then asking to generate a hexagonal stop sign) and by providing additional feedback when the generated image was incorrect, ChatGPT was sometimes able to generate the correct image as seen in **Figure 11.a.ii**. However, the ability to generate the correct image was not consistent and whenever a new chat (without previous history) was created, it failed to replicate the correct result.

We can thus conclude one more time that the LLMs we tested failed to pass even the simplest tests that follow the Turing test 2.0 framework and thus fail to showcase an intelligent level that reaches the G.I.T. We note here that various such tests can be contrived, and we have actually tested LLMs with many such test variations. In all cases LLMs failed to show the ability to consistently generate correct images so we opted to omit their results as they did not provide any new information. However, we mention some of these tests here as the reader may find it entertaining to try the tests themselves. One such test is the request to "Generate an image of a flat computer screen that has a triangular shape". Another is to "Generate an image of a car that has a total of five wheels". All such tests share the same core idea, find an image that has a typical shape/form, and attempt to generate an altered image in a somewhat rare shape (the latter is important to satisfy the third rule of the Turing test 2.0 framework). To make sure that this shape is indeed "rare enough", one can just search the web and see how easy it is to find the requested image in the requested shape. Although that does not guarantee the third rule of the Turing test 2.0 framework it is a good enough estimation as we observed from our experiments.

### 3.4 Advanced tests of the Turing Test 2.0 framework

In our previous tests, the LLMs we tested failed to demonstrate the ability to generate new *F.I.* (and thus expand their functionality) beyond their initial *F.I.* that was directly provided to them through external sources like their training data. However, as we discussed in Section 2.3, that does not necessarily mean that none of the A.I. models (the ones we tested or others) have achieved G.I. It remains an open challenge, for A.I. models to showcase that they can pass any test that follows the Turing Test 2.0 framework. If an A.I. model successfully passes any such test, then it will have reached the G.I.T. and thus achieved G.I. We emphasize that our goal in this work is not to determine if A.I. models have reached the G.I.T. but rather to provide a clear and comprehensive framework (i.e., the Turing test 2.0 framework) that can be used by any system to test if it can showcase G.I.

However, the methodology we used to generate tasks in the previous tests has some important limitations as it relies on detecting "gaps" in an A.I.'s *F.I.* and on assumptions on the nature of the training data. Such gaps in information could potentially be closed by simply balancing or expanding the training data so that LLMs can correctly solve these tests. Although in such a case, the A.I. models would still not have demonstrated G.I. (as the additional training data and would violate the second and third rules of Turing test 2.0 framework), it would make it a lot harder to construct tests that abide by the Turing test 2.0 framework rules and thus correctly detect if an A.I. has achieved G.I. To proactively solve this issue, here we propose additional tests that do not depend on assumptions on the nature of the training data used. We refer to these tests as advanced tests that use the Turing test 2.0 framework.

***Single Discipline Test***

The first advanced test we present is the Single Discipline test, where an A.I. model is trained only on a single discipline like solving math problems (science), designing logical circuits (engineering), generating images (art), generating music (art), etc. If we assume that the total *F.I.* set of that discipline today is $\boldsymbol{F'}$, then we find a subset $\boldsymbol{F} \subset \boldsymbol{F'}$ so that $\boldsymbol{I} = \boldsymbol{F'} - \boldsymbol{F}$. Then according to the Turing test 2.0 framework rules, we can say that an A.I. that was only trained to include *F.I.* set $\boldsymbol{F}$, has reached the G.I.T. if and only if it can generate any new *F.I.* information $\boldsymbol{i} \subseteq \boldsymbol{I}$.

To further explain this test let us consider an A.I. model that is designed to generate music and songs. Rather than training the model on all types and gernes of music that is known today ($\boldsymbol{F'}$), we only use data that would have been available at some predetermined time in the past e.g., the classical music period around the 1800s. Thus, we train the A.I. model only with data that represent the information set of music up to 1800s ($\boldsymbol{F}$). Note, that in this example, we empirically know that $\boldsymbol{F}$ preceded $\boldsymbol{F'}$ and that based on the available information on $\boldsymbol{F}$, future information was generated. In this case we expect the A.I. model to be able to generate music that matches classical songs, or songs before that time. However, if the A.I. model can generate music that belongs in a gerne/type of music after the 1800s ($\boldsymbol{I}$) then we can say that it has generated some information $\boldsymbol{i} \subseteq \boldsymbol{I}$ that was not available to the *F.I.* set $\boldsymbol{F}$ and thus passed the G.I.T. and achieved G.I.

Such a test does not require from the tester to detect "gaps" in the *F.I.* of a system. Instead, we start by bounding the total *F.I.* that is transmitted through training and attempt to see if the system can generate new *F.I.* Furthermore, because the system is trained in a single discipline this limits the scope of the possible functionalities that the system can gain (e.g., only need to check for tasks that generate new types of music) making it easier for the tester to select tasks that follow the Turing test 2.0 framework, and thus more reliably detect if the system has reached the G.I.T.

***Generational Test***

Although the Single Discipline test does not rely on assumptions on the nature of the training data, it has its own drawbacks. For example, it may be hard to identify what training data must be used to achieve a specific *F.I.* set. Furthermore, failing to achieve G.I. in one discipline will raise the question if the same model can achieve G.I. on a different discipline and so on. Thus, we present one more test, and probably the ultimate test that an A.I. can solve to demonstrate G.I. We call this test Generational test.

In this test an A.I. model (generation 1 model, or G1-A.I. for short) is used to train another A.I. model (G2-A.I.). That is, generated data produced by G1-A.I. (and no other data) are used as training data for G2-A.I. This process can be repeated any number of times recursively while one generation trains the next. Let us assume that G1-A.I. has a total *F.I.* set $\boldsymbol{F}$. Then G$n$-A.I. (where $n$ is any $n$-th generation that was recursively trained) has achieved G.I. if after the training is completed it can showcase a total *F.I.* set of $\boldsymbol{F'} \supset \boldsymbol{F}$, by being able to **correctly solve at least one task that the previous G1-A.I. could not**. Thus, like humans (as discussed in Section 2), if A.I. can demonstrate that it can generate more *F.I.* from one generation to another, then it has, beyond the benefit of any doubt, achieved G.I.

As a matter of fact, prior work has already tried the Generation Test against modern A.I. models, but not for the reason we propose in this work (i.e., not for testing if A.I. has achieved G.I.). Instead, prior work has attempted to use one generation of A.I. to train the next generation of A.I. in an attempt to automate A.I. training. Several such cases [31, 32, 33, 34] showcase that recursively training multiple generations of A.I. not only does not increase the information from one generation to another, but rather leads to information stagnation, also known as model collapse. After several generations, the A.I. model simply produces noise, and all *F.I.* is lost. This is a known issue in A.I. models that again shows that even today's most advanced A.I. have not yet been able to demonstrate the ability to generate new *F.I.* and thus achieving G.I. We should also note that researchers [33] proposed to include external, real data alongside the A.I. generated data when training the next generations to mitigate or completely solving the model collapse problem. However, such solutions break the fourth rule of the Turing test 2.0 framework. Thus, although such a solution helps mitigating the model collapse, it does not help with A.I. getting closer to the G.I.T.

# 4. Conclusion

We conclude this work by summarizing our contributions. First, we made a crucial distinction between the *F.I.* and *N.F.I.* that systems use to complete tasks. Based on that distinction, we were able to provide a new definition for G.I., namely the ability of a system to generate new *F.I.* This ability is necessary for systems to expand their functionality and thus necessary to achieve new advances in any discipline like science, arts, sports, etc. A system without this ability will always require an external intelligent system to provide new *F.I.* or face information stagnation. Then we defined the G.I.T. that a system must reach to claim G.I. The G.I.T. is reached if a system can demonstrate the ability to self-generate any amount of new *F.I.* by achieving any new functionality (completing a new task that previously it could not) without information input from external sources. Furthermore, we provided a novel framework called Turing test 2.0, that when followed can generate tests that can reliably test systems and decide if they reach the G.I.T. Finally, we presented several tests that follow the Turing test 2.0 and used them to test modern LLMs. Our results showed that in our proposed tests, modern LLMs are not able to demonstrate G.I. Thus, it remains an open challenge for A.I. models to use the Turing test 2.0 framework and show if they can reach the G.I.T. achieving G.I.

# 5. Related Work

Various tests have been proposed to measure the intelligence of an A.I. and investigate if the A.I. has achieved G.I. The most well-known test is the Turing test, proposed by Alan Turing [17]. The main idea of the test is to see if a computer system (e.g., A.I. model) can be distinguished from a human. A human (interrogator) interacts with two systems (test subjects) through an interface, such that the nature of the two test subjects (A.I. or human) is not revealed to the interrogator (e.g., communicating through texts using natural language). One of the test subjects is a human, where the other one is an A.I. system. The interrogator then tries to identify the nature of the test subjects and distinguish the human from the A.I. If the interrogator cannot distinguish the two test subjects the A.I. system passes the test, presuming that it has achieved G.I.

The Turing test is based on the conventional definition of G.I. as described in Section 1.1 and thus carries on the same drawbacks as the definition it relies on. Mainly there is a lot of ambiguity making such a test hard to quantify. For example, there is no clear "line of questioning" that the interrogator should use to examine the test subjects. What interactions should take place and how many are needed to reach a conclusion? Are we looking to examine specific behaviors and if yes which ones? What if an A.I. system "fools" some interrogators but not others? What if an interrogator misidentifies the two test subjects for each other? Should the human test subject be selected based on some IQ test or should any human candidate suffice? These rhetorical questions show the ambiguity and demonstrate why the Turing test, although insightful, does not suffice to help us confidently conclude if an A.I. has reached G.I.

To tackle the drawbacks of the Turing test, many other tests have been proposed (to measure A.I. intelligence), where an A.I. tries to successfully complete a specific task, that typically a human could easily complete. Goertezel et al. suggest a series of such tests [18], like testing if an A.I. can identify and operate a coffee machine inside a typical household, reading and explaining a story, passing typical school exams, learning to play a video game and others. Nilsson [19] suggested that A.I. should be able to complete specific jobs that typically humans are employed to do. Bringsjord et al. [20] suggested that A.I. should be tested using psychometric tests similar to how they are used to measure human intelligence. Additionally, a variety of benchmarks have also been proposed to more accurately measure the success rate of A.I. in completing a specific task. These benchmarks include tasks like language comprehension (reading, analyzing, and answering questions) [21, 22, 23, 25], image classification [24] and solving IQ-like tests [26].

Although all these tests give us an insight into the skills of the A.I. it is not clear to what degree they measure intelligence, if at all. For example, no good justification is given why the ability to successfully complete any of these tasks implies G.I. (other than humans can typically complete them). This is why it is important to differentiate between the ability to perform a task and demonstrating G.I. For instance, let us take the Euclidean algorithm of finding the greatest common divisor as an example. If a system can perform this task by running the Euclidean algorithm, that does not necessary imply that the system has G.I. (even though the algorithm itself points to some source of intelligence). That is why no one would argue that a computer program that runs the Euclidean algorithm (e.g., a C++ implementation of the Euclidean algorithm), has reached G.I. However, the inventor of the algorithm (in this case the ancient Greek mathematician Euclid) is the one who demonstrates G.I. by inventing the algorithm. The two (existing knowledge and adding new knowledge) are often confused.

Moreover, in the tests proposed by prior work, the "success" threshold for A.I. is set to the success rate of humans making the tests' results conditional to what humans can achieve. Additionally, the focus is always on skills that humans are better at completing compared to traditional computer systems. Instead, a test that measures intelligence and tries to detect G.I. should be unbiased against the nature of the "examinee". For example, there are plenty of tasks where machines are far better than humans, like repeating the same process multiple times with minimal deviation, recovering old memories with no information loss, high precision arithmetic calculations, etc. Why then specifically focus on skills that humans are good at? The immediate answer is that we know that humans have G.I., but that does not imply (a very common logical fallacy) that every task that humans can complete better than machines is due to their G.I.

Additionally, the above tasks are typically tasks where humans' success rate can improve after they consistently train in solving them. If the success rate of these tasks was indeed measuring intelligence, that would imply that a test subject that increases its success rate after training to solve these tasks, would have also increased their intelligence. This is a very common argument used against the accuracy of cognitive tests (a.k.a. IQ tests) in measuring intelligence, where typically one can increase their score by first studying example tests and learning how to solve similar problems. Research shows that although training can increase your score that does not imply increase in intelligence [40, 41, 42]. Thus, the success rate of such tasks depends on many other factors (other than intelligence) like prior knowledge and familiarity with the task.

For this reason, Bringsjord et al. [43] proposed restricting the total information that is used to train different A.I. systems (e.g., all systems are trained with the same training data set), and giving no advance knowledge on the specific task that A.I. will have to solve. This way, measuring intelligence can be more accurate and fairer when comparing different systems. Bringsjord et al. also propose the Lovelace test [44] where an A.I. system is considered to have achieved G.I. only if we can prove that the A.I. has produced a result, such that the information needed for that result does not originate from the information used to program and train the A.I. In other words, A.I. can produce original, new ideas. Although this threshold of G.I. (producing original new ideas) may sound similar to our definition of G.I.T., it has some crucial differences. In our work, the new ideas (which we refer to as *F.I.*) directly come from information in the system. However, the latter information is what we call *N.F.I.* That is, information that although is contained in the system it does not directly contribute to the system's functionality, until the system is able to generate understanding and meaning from it.

Another interesting definition for G.I. comes from Chollet in [26], where it is defined as the ability to gain new skills. Although the two definitions between [26] and [44] may sound similar, the main difference is that in [26] the new skills can be acquired through training, where humans act as an external source of information by transmitting new *F.I.* to a system as we discussed in Section 2.1. Contrary in [44] the A.I. is expected to "surprise" its creators by producing an unexpected result, like acquiring a skill that the designers/testers of the A.I. did not design/train the A.I. for.

## 6. Acknowledgements

The author thanks everyone that provided feedback and helped with revising this work. Special thanks go to C. Rossides for his extensive feedback and comments that strengthened this work. This work did not receive any specific grant or financial support from public, commercial, or not-for-profit funding agencies.